\documentclass[a4paper,twoside]{article}

\usepackage{graphicx}
\usepackage{epsfig}
\usepackage{subcaption}
\usepackage{calc}
\usepackage{amssymb}
\usepackage{amstext}
\usepackage{amsmath}
\usepackage{amsthm}
\usepackage{multicol}
\usepackage{pslatex}
\usepackage{apalike}
\usepackage{algorithm2e}
\usepackage[bottom]{footmisc}
\usepackage{fancyhdr} 
\usepackage{xcolor} 
\usepackage{SCITEPRESSpagenumbers}     

\fancypagestyle{firstpage}
{
    \fancyhead[L]{\color{red} To be presented/published at the 16th International Conference on Agents and Artificial Intelligence (ICAART2024); Rome, Italy, 24-26 February 2024.}    
    \fancyhead[R]{\color{blue} Copyright \copyright 2024 C.\ Terawong \& D.\ Cliff}
}

\begin{document}

\thispagestyle{firstpage}

\title{XGBoost Learning of Dynamic Wager Placement for In-Play Betting on an Agent-Based Model of a Sports Betting Exchange}

\author{\authorname{Chawin Terawong and Dave Cliff\orcidAuthor{0000-0003-3822-9364}}
\affiliation{Department of Computer Science, University of Bristol, Bristol BS8 1UB, U.K.}
\email{\{ae22586, csdtc\}@bristol.ac.uk}
}


\abstract{We present first results from the use of XGBoost, a highly effective machine learning (ML) method, within the {\em Bristol Betting Exchange}
(BBE), an open-source agent-based model (ABM) designed to simulate a contemporary sports-betting exchange with in-play betting during track-racing events such as horse races. We use the BBE ABM and its array of minimally-simple bettor-agents as a synthetic data generator which feeds into our XGBoost ML system, with the intention that XGBoost discovers profitable dynamic betting strategies by learning from the more profitable bets made by the BBE bettor-agents. After this XGBoost training, which results in one or more decision trees, a bettor-agent with a betting strategy determined by the XGBoost-learned decision tree(s) is added to the BBE ABM and made to bet on a sequence of races under various conditions and betting-market scenarios, with profitability serving as the primary metric of comparison and evaluation. Our initial findings presented here show that XGBoost trained in this way can indeed learn profitable betting strategies, and can generalise to learn strategies that outperform each of the set of strategies used for creation of the training data. To  foster further research and enhancements, the complete version of our extended BBE, including the
XGBoost integration, has been made freely available as an open-source release on GitHub.}

\onecolumn \maketitle \normalsize \setcounter{footnote}{0} \vfill

\section{\uppercase{Introduction}}
\label{sec:introduction}

Like many other long-standing aspects of human culture, despite its five-thousand-year history, gambling activity and opportunities were transformed by the rise of the World-Wide-Web in the dot-com boom of the late 1990s. One particular technology innovation from that time subsequently proved to be a seismic shift within the gambling industry: this was the arrival of commercial web-based {\em betting exchanges}. 

In much the same way that financial markets such as stock exchanges offer platforms where potential buyers and potential sellers of a stock can interact to buy and sell shares, with buyers and sellers indicating their intended prices in {\em bid} and {\em ask} orders, which are then matched to compatible counterparties by the exchange's internal mechanisms, so betting exchanges are platforms where potential {\em backers} and potential {\em layers} can interact and be matched by the exchange, to find one or more people to take the other side of a bet. In the terminology of betting markets, a backer is someone who places a {\em back} bet, i.e.\ a bet which will be paid if a specific event-outcome does occur; and a layer is someone who places a {\em lay} bet,  i.e.\ a bet that's paid if the specific event-outcome does {\em not} occur. The revolutionary aspect of betting exchanges is that they operate as platform businesses: the exchange does not take a position as either a layer or a backer, it simply serves to match customers who want to back at a particular odds with other customers who want to lay at those same odds, and the exchange makes its money by taking a small fee from each customer's winnings. In contrast, traditional bookmakers (or ``bookies'') are the counterparty to each customer's bet, and lose money if they miscalculate their odds.

The first notably commercially successful sports betting exchange was created by British company BetFair (see {\tt www.betfair.com}), a start-up which grew with explosive pace after its founding in 2000, and by 2006 was valued at \pounds1.5billion. In 2016 Betfair merged with another gambling company, Paddy Power, in a deal worth £5bn, and the Betfair-branded component of the merged company (now known as Flutter Entertainment {\sc plc}) remains the world's largest online betting exchange to this day; at the time of writing this paper in late 2023, Flutter's market capitalization is \pounds22.5billion. For further discussion of BetFair, see e.g. \cite{davies_etal_2005,houghton_2006,cameron_2009}

Creating an online exchange for matching layer and backer bettors was not the only innovation that BetFair introduced. They also led in the development of {\em in-play} betting, which allowed bettors to continue to place back and lay bets after a sports event had started, and to continue betting as the event progressed, until some pre-specified cut-off time or situation occurred, or the event finished. This is in contrast to conventional human-operated bookmakers, who ceased to take any further bets once the event of interest was underway: because Betfair's betting exchange system was entirely automated, it could process large numbers of bets while an event is underway, operating in real time  with flows of information that would overwhelm a human bookie. 

Just as most stock-exchanges publish real-time summary data of all the bids and asks currently seeking a counterparty, often showing the quantity available to be bought or sold at each potential price for a particular stock, so a betting exchange publishes real-time summary data for any one event $E$ showing all the currently unmatched backs and lays, the odds (or ``price'') for each of them, and the amount of money available to be wagered at each price -- in the terminology of betting exchanges, this collection of data is the ``market'' for event $E$. 

During in-play betting, the prices in the market can shift rapidly, and while some types of events such as tennis matches might last for hours, allowing for hours of in-play betting to endure for a single match, for other types of event such as horse-racing the event may only last a few minutes. The exploratory work that we describe in this paper is motivated by the hypothesis that it may be possible to use machine learning (ML) methods to process the rapidly-changing data on a betting-exchange market for short-duration events such as horse races, and for the ML system to thereby produce  novel profitable betting strategies. 

For the rest of this paper, without loss of generality, we'll limit our descriptions to talking only of betting on horse races because this is a very widely known form of sport on which much money is wagered, because the duration of most horse races is only a few minutes, and also because it is reasonably easy to create an appropriately realistic agent-based model (ABM), a simulation of a horse race, where each agent in the model represents a horse/rider combination, and where during the race each agent has a particular position on the track, is travelling at a specific velocity, and may or may not be blocked or otherwise influenced by other horses in the race. Exactly such a simulation of a horse race was introduced by Cliff \cite{cliff_2021_bbe}, as one component of the {\em Bristol Betting Exchange} (BBE), an agent-based simulator not only of horse races, but also of a betting exchange, and also of a population of bettor-agents who each form their own private opinion of the outcome of a race, and then place back or lay bets accordingly. Various implementations of BBE have been reported previously by \cite{cliff_etal_2021_emss} and at ICAART2022 by \cite{guzelyte_cliff_2022}, but to the best of our knowledge ours is the first study to explore use of XGBoost \cite{chen_guestrin_2016} on in-play betting data to develop profitable betting strategies. 

The bettor-agents in the BBE ABM each form their own private opinion on the outcome of a race on the basis of their own internal logic, i.e.\ their own individual betting strategy, and the original specification of BBE in \cite{cliff_2021_bbe} included a number of minimally simple strategies, described in more detail in Section~\ref{sec:bbe_bettors} below, and the BBE ABM usually operates with a bettor population having a heterogeneous mix of such betting strategies. As the dynamics of a simulated race unfold, so the bettor-agents react to changes in the competitors's pace and relative positions by making and/or cancelling in-play bets, altering the market for that race. The BBE ABM records every change in the market over the duration of a race, along with the rank-order positions of the competitors at the time of each market change (i.e., which competitor is in first place, which is in second, and so on): this we refer to as a {\em race record}.  

In the work described here, we typically run 1000 race simulations, gathering a race record from each. The set of race records then go through an automated analysis process to identify the actions of the most profitable bettors in each race. That is, for each race, we look to see which bettors made the most profit from in-play betting on that race, and we then work backward in time to see what actions those bettors took during the race, and what the state of the market and the state of the race was at the time of each such action. This then forms the training and/or test data for XGBoost: for any one item of such data, the input to XGBoost is the state of the market and the state of the race, and the desired output is the action that the bettor took.

To accomplish this, we modified the existing source-code of the most recent version of BBE, which
is the multi-threaded BBE integrated with Opinion Dynamics Platform used in Guzelyte’s research \cite{guzelyte_2021,guzelyte_cliff_2022},
hosted on Guzelyte’s GitHub \cite{guzelyte_2021_github}, to incorporate the XGBoost betting agent. After the integration of XGBoost, we conduct experiments to train and evaluate the XGBoost bettor-agent’s performance in different market scenarios. 

A surprising result we present here is that although XGBoost is trained on the profitable betting activity of a population of minimally simple betting strategies, the betting strategy that it then learns can outperform even the best of those simple strategies. That is, XGBoost generalises over the training data sufficiently well that the profitability of XGBoost-enabled bettor-agents can eventually be better than those of the non-learning bettors whose behaviors formed the training data for XGBoost.  

Our work described here is exploratory: we use the BBE agent-based model (ABM) as a synthetic data generator to create the training data needed for XGBoost, and then we take the XGBoost-trained bettor agent and test it in the BBE ABM. We are doing this in an attempt to answer the research question of whether the multidimensional time series of data from the in-play betting market for horse races can in principle be fed into a machine learning system such as XGBoost and result in a learned profitable automated betting system. What we develop here is a proof-of-concept, and as we show in Section~\ref{sec:results} our results thus far do show some promise, but we strongly caution against any readers of this paper actually gambling with real money on the basis of the system we describe here: a lot of further development work and much more extensive testing would be required before we would ever want to deploy this system live-betting with our own money.

Section~\ref{sec:background} gives further background information, and then our experiment design is described in Section~\ref{sec:expdesign}. Section~\ref{sec:results} presents our results, followed by discussion of future work in Section~\ref{sec:future}.

\section{Background}
\label{sec:background}

\subsection{BBE Race simulator}

Guzelyte’s research \cite{guzelyte_2021,guzelyte_cliff_2022}, relies heavily on Keen’s thesis \cite{keen_2021_MEng} and Cliff’s original paper \cite{cliff_2021_bbe} to create the
Bristol Betting Exchange (BBE) race-simulator. BBE isn’t designed to perfectly mimic a real track
horse-race, but rather to generate convincing data resembling real race dynamics: the changes in pace and rank-order position that occur within real races. 

In every race simulation, competitors are selected from a pool and placed on a one-dimensional track
of a fixed length. Each competitor’s position on the track at a given time is represented
as a real-valued distance. The race begins at $t=0$ and concludes when the last competitor crosses
the finish line.  The progress of each competitor is calculated using a discrete-time stochastic process which provides for modelling of individual differences in pace (e.g., some might start the race at a fast pace but subsequently slow down; others might instead hold back in the early stages of the race and then speed up at the end) and for inter-competitor interactions such as a competitor being blocked and hence slowed by another competitor immediately in front, or a competitor being ``hurried'' by another competitor closing on to its rear. Full details of the race simulator 
were first published in \cite{cliff_2021_bbe} to which the reader is referred for more information

\subsection{BBE Betting Exchange}

The Bristol Betting Exchange (BBE) uses a matching-engine that tracks all of the bets that are placed.
The details of the bets of each bettor (time of bet, amount of bet, etc) is recorded for every race. If
a bet hasn’t been matched with a counterparty, it can be cancelled, but once it’s matched, it can’t be.
When it comes to matching bets, older bets are prioritized if the odds are the same. To create the market for a race, for each competitor BBE collects all back
and lay bets at the same odds, calculating the total money bet. After a race finishes, the money from
losing bets is gathered and given to the winners. BBE earns by taking a small commission from the
winnings. BBE’s matching-engine is designed to implement exactly the same processes as are used in real-world betting exchanges, so in this sense it can
be argued that the BBE’s exchange module is not just a simulation, but an actual instance of a betting exchange.

\subsection{BBE Betting Agents}
\label{sec:bbe_bettors}

BBE has a variety of bettor-agents each utilizing a unique approach.
These bettors’ strategies range from the wholly irrational {\em Zero Intelligence} (ZI) strategy, where choice of bets is made entirely at random; to wholly rational, where the best available information guides their decisions. The most rational strategy in BBE at present is the  {\em Rational Predictor} (RP), which makes its race outcome predictions based on a series of simulated
``dry-runs'' of the race: at the start of the race, an RP bettor runs $n$ independent and identically distributed (IID) private (i.e., known only to that RP bettor) simulations of the entire race from time $t=0$ to whatever time the last horse crosses the finish-line, using the same race simulator engine as is used to implement the actual `public' race that all BBE bettors are betting on; then at various times $t=t_i$ during the race, the RP bettor will run another fresh set of $n$ IID simulations of the current race, running the simulation forwards from the {\em current} positions of the horses at time $t_i$ forward to the end of the race, and then may then make a fresh in-play bet if the most frequent winning horse in those $n$ simulations is different from whatever horse it had previously bet on.  In \cite{cliff_2021_bbe}, the behavior of an RP agent was defined as being determined primarily by the hyperparameter $n$, how many IID simulations it runs each time it reevaluates the odds, and so these agents were denoted as RP($n$) bettors. Other authors who have worked with BBE since the publication of \cite{cliff_2021_bbe} have found it useful to also be explicit about the time interval between an RP($n$) bettor's successive sets of $n$ IID simulations: this can be captured by two additional hyperparameters, $\Delta_{t\text{:min}}$ and $\Delta_{t\text{:max}}$, such that the wait (in seconds) until the next set of $n$ IID simulations is conducted by an RP bettor is given by a fresh draw from a uniform distribution between $\Delta_{t\text{:min}}$ and $\Delta_{t\text{:max}}$ (denoted by ${\cal U}[\Delta_{t\text{:min}}, \Delta_{t\text{:max}}]$) as that bettor concludes its current set of IID simulations. For this reason, an RP bettor is fully denoted by RP($n,\Delta_{t\text{:min}}, \Delta_{t\text{:max}}$). 

The computational costs of simulating any one RP($n,\Delta_{t\text{:min}}, \Delta_{t\text{:max}}$) bettor agent over the duration of an entire race manifestly rises sharply as $n$ increases and/or as the expected value $E({\cal U}[\Delta_{t\text{:min}}, \Delta_{t\text{:max}}])$ falls.  Authors such as  \cite{keen_2021_MEng,guzelyte_2021} have concentrated on using the relatively low-computational-cost instance of RP(1,10,15), which --- because this type of bettor is in receipt of privileged information --- has come to be referred to as the {\em Privileged} bettor strategy. In the work presented here, we follow their convention and also use Privileged bettors as our form of RP agent.

There are several other types of BBE bettor strategies. The {\em Linear Extrapolator} (LinEx) employs a
strategy of estimating competitor speed and predicting the outcome based on linear extrapolation. The
{\em Leader Wins} (LW) bettor operates on the assumption that the leading competitor will maintain their
position and win the race. The {\em Underdog} strategy (UD) supports the second-placed competitor as long as they
are within a certain threshold of the lead. The {\em Back The Favourite} (BTF) bettor, on the other hand,
aligns their predictions with the market’s favourite.

The {\em Representative Bettor} (RB) is a unique agent designed to mimic real-world human betting behaviours.
It factors in betting preferences such as an inclination towards certain stake amounts, often seen
in human bettors who prefer multiples of 2, 5, or 10. This bettor also exhibits the well-known {\em favorite-longshot bias}, reflecting the tendencies of human bettors to bet disproportionately on the favourite or
the longshot, regardless of the actual odds.

The presence of these various bettors, each with different parameters, within BBE gives
rise to an engaging and complex dynamic in the in-play betting market. For full details and
implementation notes on each of these bettor strategies, see \cite{cliff_2021_bbe,keen_2021_MEng,guzelyte_cliff_2022}. 

\subsection{XGBoost Model Training}

XGBoost, an abbreviation for eXtreme Gradient Boosting, introduced by \cite{chen_guestrin_2016}, is an ML technique
celebrated for its speed, precision, and flexibility. The algorithm operates on the gradient boosting
framework, sequentially crafting decision trees that progressively enhance prediction accuracy.
XGBoost has proven its effectiveness by frequently being used in winning solutions to international data science competitions, particularly on Kaggle, a competitive data science platform \cite{adebayo_2020}. Moreover, its real-world use extends to different fields like
predictive modelling and recommendation systems, demonstrating its versatility. With its combination
of computational power and predictive capability, XGBoost continues to drive advancements in machine
learning.

Gradient boosting is a technique utilized in machine learning for both regression and classification problems.
It operates by iteratively combining a series of simple predictive models, typically decision trees.
Each subsequent model is designed to rectify the residual errors of its predecessor, thus enhancing accuracy
incrementally \cite{Friedman_2001}.

The technique gets its name from Gradient Descent, an optimization method used to minimize the
chosen loss function. Every new model reduces the loss by moving in the direction of the steepest descent.
It does this by incorporating a new tree that minimizes the loss most effectively, which is the essence
of gradient boosting \cite{Friedman_2001}. Unlike other boosting algorithms such as AdaBoost \cite{freund_schapire_1997}, gradient boosting identifies
the weaknesses of weak learners via gradients in the loss function, while AdaBoost does so by examining
high-weight data points.

XGBoost \cite{chen_guestrin_2016} is designed to be highly scalable and parallelizable, making it suitable for handling large-scale
datasets. The algorithm effectively manages sparse data and missing values without additional input.
Additionally, XGBoost includes a regularization term into its objective function. This regularization term
helps control the model complexity and avoid overfitting, a common problem found with other gradient
boosting algorithm \cite{Friedman_2001}.

Space limitations prevent us from giving here full details of the XGBoost algorithm, for which the reader is instead referred to \cite{chen_guestrin_2016}. For the purposes of this paper, it is sufficient to treat XGBoost as a ``black-box'' learning method that produces a set of decision trees that can be used as a betting strategy.  We used the XGBoost Python library \cite{chen_2023} which provides a flexible interface with the Scikit-learn API \cite{pedregosa_etal_2011}.

XGBoost’s performance can be further improved with a technique called parameter tuning. The
algorithm’s parameters and hyperparameters are divided into three types: general parameters, booster
parameters, and learning task parameters \cite{chen_2023}. General parameters control the overall function, booster
parameters influence individual boosters, and learning task parameters oversee the optimization process.
Example of hyperparameters include: maximum tree depth ({\tt max\_depth}); step size shrinkage ({\tt learning\_rate}),
number of trees ({\tt n\_estimators}); minimum loss reduction ({\tt gamma}); minimum sum of instance weight
({\tt min\_child\_weight}); and the subsample ratio of training instances ({\tt subsample}). Through the adjustment
and tuning of these parameters, XGBoost can be adjusted to efficiently address a broad range of
machine learning tasks.

\subsection{Tuning and Cross Validation}

{\bf Hyperparameter Tuning using Grid Search:}
In machine learning, hyperparameters are important configurations that pre-determine the algorithm’s
training process which influences the model’s final performance on prediction accuracy \cite{nyuytiymbiy_2020}. In particular,
within the scope of this research involving the XGBoost algorithm, key hyperparameters such as the
learning rate and the maximum depth of the decision trees can significantly affect the model performance.
The optimization of these hyperparameters is core to achieving the highest model prediction ability.
A widely recognized technique to find the best set of hyperparameters is called ‘Hyperparameter Tuning’.
Among the variety of methods available for this purpose, ‘Grid Search’ emerges as particularly robust.
Grid Search conducts a methodical exploration of all potential combinations of hyperparameter values
within a predefined boundary (list). This is for finding the combination of hyperparameters that offers
optimal model performance \cite{malato_2021}.

{\bf Cross Validation (K-Fold cross-validation):}
K-fold cross-validation is a technique to determine the performance of a machine learning model. It
involves partitioning the dataset into $k$ equally-sized subsamples. Each iteration uses one subsample for
validation and the remaining $k-1$ for training. The model undergoes $k$ evaluations, with each subsample
serving once as the validation set. The outcomes from the $k$ tests are averaged to obtain a consolidated
performance metric. This approach ensures every data point contributes to both training and validation,
preventing overfitting of the model \cite{scikit_learn_xval}.

{\bf GridSearchCV:}
The GridSearchCV module from the Scikit-learn library \cite{scikit_learn_gridsearch} integrates grid search and cross-validation,
offering a streamlined mechanism for hyperparameter tuning. To use it, one provides a model (like
XGBoost), a parameter grid defining the hyperparameter value combinations to test, a scoring method, and a
number of k-fold cross-validations. GridSearchCV then systematically explores these combinations using
cross-validation, where the dataset is partitioned into subsets and each subset is iteratively used for
validation. Leveraging GridSearchCV with the XGBoost algorithm not only saves effort but also ensures the identification of optimal hyperparameters, enhancing model robustness
and performance on unseen data.

\section{\uppercase{Experiment Design}}
\label{sec:expdesign}

\subsection{Overview}

The primary objective of this experiment is to integrate the XGBoost betting agent into the existing
suite of agents in the BBE system. The agent will leverage a trained XGBoost model to make informed
decisions on whether to ‘Back’ or ‘Lay’ a bet, predicated on the input data.

\begin{figure*}
\centering
\includegraphics[width=0.8\linewidth]{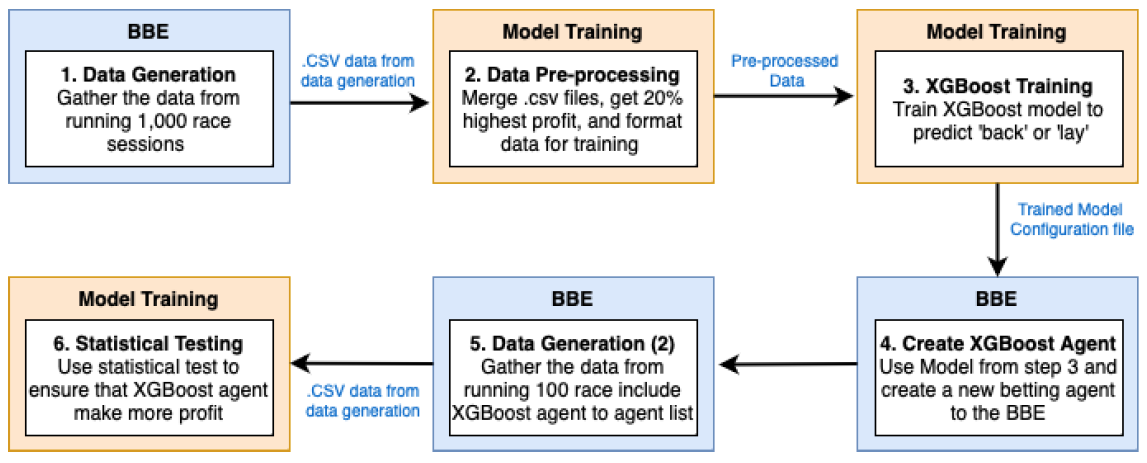}     
\caption{High-level overview of the experiment and the data flow of the system.}
\label{fig:3.1}
\end{figure*}

Figure~\ref{fig:3.1} provides the high-level design of this experiment. The first step is to gather the data from BBE by running 1,000 race sessions. These race records are then pre-processed, narrowing down the data to the top 20 percent of
the most significant transactions (Back or Lay actions). This refined dataset was then used for training
with the XGBoost Python library. By tuning the Hyperparameters and using Cross-validation, the goal
was to ensure the model could perform well in many scenarios.

Once a trained XGBoost model is ready, it is added as a new betting agent into the BBE system. To
further test its performance, various race scenarios were run, each scenario with 100 races. This approach
ensured we had enough data to assess how the new agent compared to the existing ones. Lastly, the
collected data is used for statistical hypothesis tests to validate if the new XGBoost agent is more
profitable.

\subsection{Setup for Model Training}

Here we outline the specific scenario used for the data gathering process for XGBoost model training. Complete further details  details are available in \cite{terawong_2023}. We ran 1000 races, each of the same fixed distance so that the duration of each race was approximately the same, and all races had 5 competitors. The population of bettors in the ABM was made up from 10 each of Guzelyte's \cite{guzelyte_2021,guzelyte_cliff_2022} ``opinionated'' versions of the ZI, LW, BTF, LinEx, and Underdog strategies plus 5 of Guzelyte's Opinionated-Privileged strategy; and then 55 of the original un-opinionated versions of these strategies, again split 10/10/10/10/10/5, for a grand total of 110 bettors.

\subsection{Setup for Profit Validation}
\label{sec:3.1.4}

After implementing the XGBoost agent into the BBE system, data (including the data generated by the
XGBoost agent) was gathered to evaluate whether the XGBoost agent generates more profit than the
other agents. Two scenarios are created to experiment with this new XGBoost agent.

\begin{itemize}

\item
{\bf Scenario 1:} The number of simulations is reduced from 1,000 to 100, as this is only for profit testing
and not for model training. A total of 5 XGBoost-trained bettor agents were added to the population. 

\item
{\bf Scenario 2:} Everything remains the same as in Scenario 1, except the number of agents is set to 5
for every agent type. This is for testing how the XGboost agent performs when the environment
changes. 

\end{itemize}

\subsection{XGBoost Parameters}

We used the Scikit-learn XGBoost API instead of the native XGBoost API. The
native API of XGBoost provides a highly flexible and efficient way to train models, making it suitable
for those experiment that prioritize performance and more refined configuration. On the other hand, the
Scikit-learn XGBoost API is a wrapper around this native API that integrates seamlessly with the widely
used Scikit-learn Python Library. This compatibility is the primary reason for selecting it in this research,
mainly due to its seamless connection with GridSearchCV. This tool aids in hyperparameter tuning, an
important aspect of model optimization. Moreover, the Scikit-learn XGBoost API offers a user-friendly
interface that reduces some complexities while still retaining robustness and enough flexibility.

In the model training process, specific choices shaped its direction. One crucial decision was the
selection of the model’s objective function. This function dictates what the model aims to achieve during
the learning process. In this work reported here the objective chosen was {\em binary:logistic}. This objective means
the model is made to perform binary classification, determining an output as one of two distinct classes.
The term logistic refers to the logistic function, mapping any input
into a value between 0 and 1, making it suitable for probability estimation in binary decisions. Given
the context of our betting decisions being binary (Back or Lay), the binary:logistic objective was a
proper selection \cite{chen_2023}.

For evaluating the effectiveness of the model, logistic loss (``LogLoss'') was chosen as the metric \cite{chen_2023}.
In machine learning, the choice of evaluation metric is crucial as it directly influences how the model’s
performance is estimated and how it learns during training. The LogLoss is a measure for binary
classification that quantifies the accuracy of a classifier by penalizing false classifications. It estimates
the probabilities associated with the accuracy of predictions. The closer the predicted probability is to
the actual class, the lower the log loss.

Choice of hyperparameter values is crucial in shaping model performance. Hyperparameter tuning played a crucial
role in the training, and we used GridSearchCV. The parameters tuned in our work are as follows: the learning rate {\tt eta} which regulates step-size during boosting; {\tt max\_depth}, the maximum depth of the decision tree; {\tt subsample}, the fraction of training data to be used in each boosting round; {\tt colsample\_bytree}, the fraction of features to be used for constructing each tree; {\tt gamma}, the minimum loss reduction required to make a further partition of a leaf node in the decision tree, and {\tt n\_estimators}, which indicates the number of boosting rounds or trees
to be constructed. This sequence is followed because n estimators influences training time. Identifying the
best hyperparameter set initially reduces the training time on finding the appropriate n estimators.

For further details of the design and implementation of this extended version of BBE, see \cite{terawong_2023}.

\section{\uppercase{Results}}
\label{sec:results}

\subsection{Evaluation of XGBoost Training}

\subsubsection{Evaluation of Hyperparameters}

\begin{figure}
\centering
\includegraphics[trim={15cm 0cm 15cm 1cm},clip,width=0.6\linewidth]{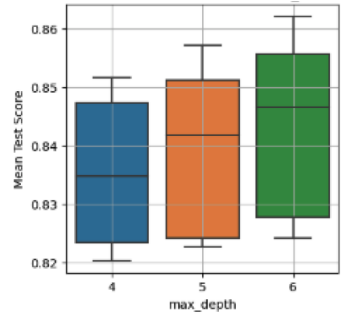}     
\caption{Box-plot of the influence of XGBoost hyperparameter  {\tt max\_depth} on the mean test score.}
\label{fig:4.1maxdepth}
\end{figure}

\begin{figure}
\centering
\includegraphics[trim={30cm 0.5cm 0cm 1.5cm},clip,width=0.65\linewidth]{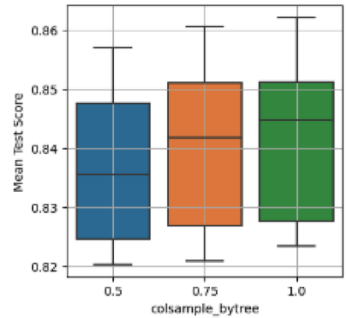}     
\caption{Box-plot of the influence of XGBoost hyperparameter  {\tt colsample\_bytree} on the mean test score.}
\label{fig:4.1colsample}
\end{figure}

\begin{figure}
  \centering
\includegraphics[width=0.99\linewidth]{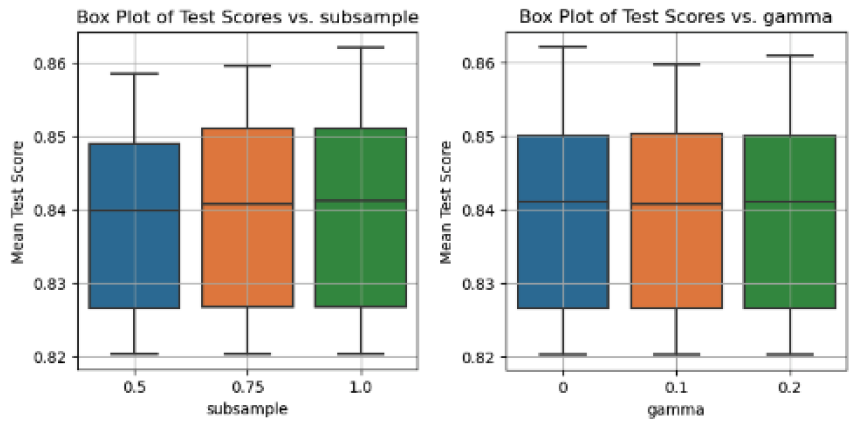}    
  \caption{Box-plots illustrating the influence of {\tt subsample} (left) and {\tt gamma} (right)  on the mean test score.}
  \label{fig:4.2}
 \end{figure}

The metric for evaluating hyperparameter performance in the experiment is the
‘accuracy’ score of the classification. Figures~\ref{fig:4.1eta} to ~\ref{fig:4.2} present results from a 5-fold cross-validation combined with a hyperparameter tuning using grid search on the training dataset for the XGBoost model. Key insights are summarized as follows:

\begin{figure}
\centering
\includegraphics[trim={0cm 0cm 30cm 1cm},clip,width=0.6\linewidth]{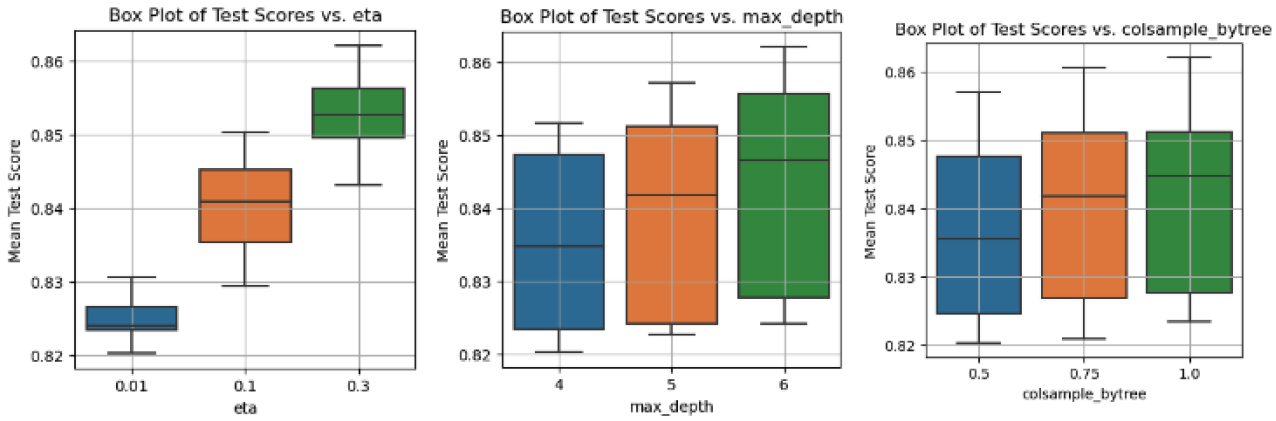}     
\caption{Box-plot of the influence of XGBoost hyperparameter {\tt eta} on the mean test score.}
\label{fig:4.1eta}
\end{figure}

\begin{itemize}

\item {\bf Hyperparameters impacting model performance:} As shown in Figures~\ref{fig:4.1eta} to~\ref{fig:4.1colsample}, the hyperparameters
 {\tt eta}, {\tt max\_depth}, and {\tt colsample\_bytree} significantly influence model performance. An
increase in the values of these hyperparameters generally correlates with an improved accuracy. Among them, {\tt eta} demonstrates a pronounced effect, whereas {\tt colsample\_bytree}
exhibits a more subtle impact.

\item {\bf Hyperparameters with minimal impact:} Variations in hyperparameters like {\tt subsample} and
{\tt gamma} seem to have little to no effect on the model’s performance according to Figure~\ref{fig:4.2}.

\end{itemize}

Through the process of cross-validation and hyperparameter tuning on the training data using the
XGBoost machine learning algorithm, an optimal model was derived with a specific set of hyperparameters: {\tt colsample\_bytree}=1.0; {\tt eta=0.3}; {\tt gamma}=0; {\tt max\_depth}=6; and {\tt subsample}=1.0.

\subsubsection{Flow of Control}

After identifying the optimal set of learning hyperparameter values, the next two hyperparameters to determine concern the flow of control in XGBoost. The hyperparameter
{\tt n\_estimators} indicates the number of boosting rounds or trees that should be constructed, and we set this to 1,000.
We also used {\tt early\_stopping\_rounds}=10 which ensures that training halts if the validation
metric doesn’t demonstrate any enhancement over 10 successive boosting rounds. The combined
usage of {\tt n\_estimators} and {\tt early\_stopping\_rounds} aids in mitigating both underfitting and overfitting
of the model.

\begin{figure} 
\centering
\includegraphics[trim={2mm 1mm 12mm 0mm},clip,width=0.99\linewidth]{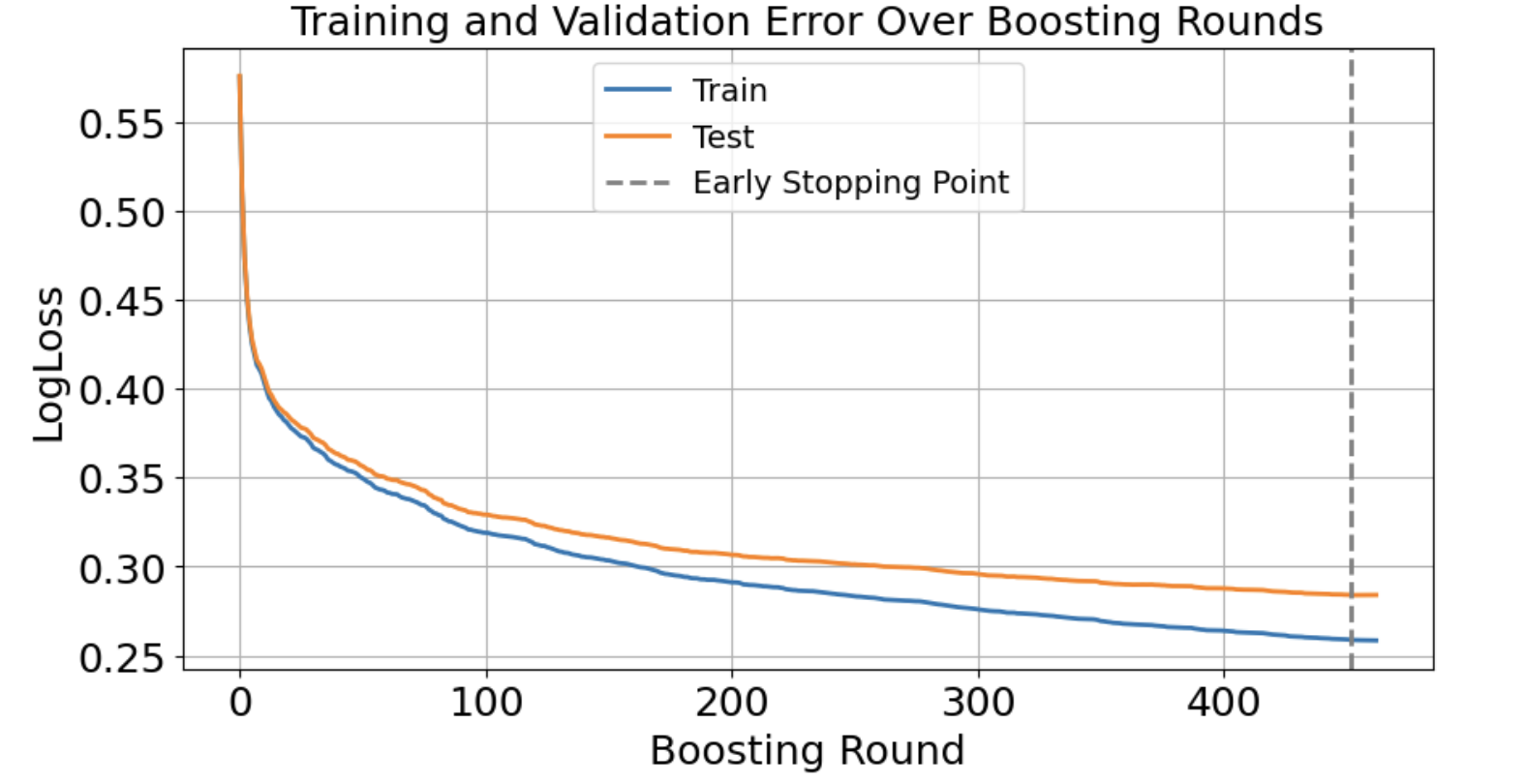}     
\caption{Illustration of the learning process of the model. It can be observed that the LogLoss
gradually decreased when the number of boosting round increases.}
\label{fig:4.3}
\end{figure}

As depicted in Figure~\ref{fig:4.3}, the LogLoss consistently decreases as the number of boosting rounds
increases. This suggests that the model continues to learn and improve. The dashed line represents
the early stopping point, beyond which the model no longer exhibits significant improvement. The
model halted at the 452nd boosting round, as indicated by the early stopping mechanism. With these
refinements, an optimal model has been obtained.

\subsubsection{Evaluation of Optimal XGBoost Model}

The SciKitLearn XGBoost training function that we use in this work provides insights into how different features contribute to the model’s predictions using the XGBoost F-score, which denotes the frequency with which a feature is
used to split the data across all trees: this tell us the relative importance of each feature. The three most important features in our XGBoost bettor were:

\begin{enumerate}

\item Distance (F-score: 8680): As the most influential feature, ‘distance’ plays a central role in the
model’s decision-making, indicating its significance in predicting the patterns the model identifies.

\item Time (F-score: 6642): The ‘time’ feature emerges as the second most influential feature, though
it lags behind ‘distance’ by over 2000 points. Nevertheless, its considerable F-score denotes its
relevance in the model’s predictions.

\item Rank (F-score: 1276): With a considerably lower F-score compared to the other two features,
‘rank’ seems to have a more marginal impact on the model’s decision process.

\end{enumerate}

The confusion matrix of Table~\ref{tab:confusionmatrix} summarizes the performance of the classification
model. The following interpretations can be drawn:

\begin{table}
\caption{Confusion Matrix from the XGBoost training}
\label{tab:confusionmatrix} \centering
\begin{tabular}{r|c|c|}
   & \multicolumn{2}{c}{Predicted Labels}  \vline \\
  \hline
  & lay & back \\ 
  \hline
  True Labels & &   \\
  \hline
  lay & 72521 & 1197  \\
  back & 9541 &  6730 \\
  \hline
\end{tabular}
\end{table}

\begin{itemize}

\item True Positive (6730): These instances correctly identify a back bet. This means the model
correctly predicted 6,730 instances where one should back a bet.

\item True Negative (72521): These instances represent cases where the model correctly predicted
a lay bet. In this context, it means the model accurately identified 72,521 instances where one
should not back the bet.

\item False Positive (1197): These instances represent errors in prediction. The model mistakenly
identified 1,197 bets as back when they should have been lay.

\item False Negative (9541): This count represents instances where the model wrongly classified bets
as lay when they should have been back. These could be seen as signifying areas where the model
could be improved for better accuracy since the number is high.

\end{itemize}

According to this matrix, the model shows high accuracy in predicting when to place a lay bet because the
number of true negatives is high. However, the number of false negatives indicates an area of potential improvement 
for the model, highlighting the potential to better recognise instances when the bettor should place a back  bet.

The classification report, summarized in Table~\ref{tab:4.2}, provides in-depth metrics to assess the XGBoost
model’s performance on the betting decision. Here, Class 0 corresponds to {\em lay} and Class 1 to {\em back}:

{\bf Precision} (Class 0: 0.88; Class 1: 0.85): Precision measures how accurate the model’s positive
predictions are. For Class 0, a precision of 0.88 means that out of all predicted lays, 88\%
were correct. For Class 1, 85\% of the model’s predictions were correct.

{\bf Recall} (Class 0: 0.98; Class 1: 0.41): Recall assesses the model’s ability to detect all actual
positives. For Class 0, the model identified 98\% correctly. However, for Class 1, it
was only 41\%, showing room for improvement here.

{\bf F1 Score} (Class 0: 0.93; Class 1: 0.56): F1-Score balances precision and recall. While it shows
a high score for Class 0 of 93\%, it has a moderate score for Class 1 of 56\%.

{\bf Accuracy} (0.88): The model correctly predicted the outcome for 88\% of the bets.
Referring to Table~\ref{tab:4.2}, it’s clear that the model performs well for Class 0 because precision and recall
are high. For Class 1, while precision remains high, recall drops, indicating challenges in detecting this
class. However, the model’s overall prediction accuracy is still high at 88\%.

\begin{table}[h]
\caption{Classification Report of the best model trained by XGBoost: Class 0 is lays; Class 1 is backs; {\em Acc}.\ is Accuracy; {\em MA} is Macro Average; and {\em WA} is Weighted Average}
\label{tab:4.2} \centering
\begin{tabular}{|c|c|c|c|c|}
  \hline
  Class & Precision  & Recall &  F1 & Support \\
  \hline
  0 & 0.88 & 0.98 &  0.93 & 73718 \\
  1 & 0.85 &  0.41 & 0.56 & 16271 \\
Acc. & & & 0.88 & 89989 \\
MA & 0.87 & 0.70 & 0.74 & 89989 \\
WA & 0.88 & 0.88 & 0.86 & 89989 \\
  \hline
\end{tabular}
\end{table}

\subsection{Hypothesis Testing}
\subsubsection{Scenario 1}

{\bf Simulation Setup:}
The sessions were set at 100 rounds, incorporating various agents in predefined quantities. Specifically,
10 agents each of types ZI, LW, Underdog, BTF, and LinEx were introduced,
along with 5 each of XGBoost and Privileged agents. This specific configuration was reminiscent of the
one used during the model training phase, albeit here with the inclusion of the XGBoost agent. 

\begin{figure*}
\centering
\includegraphics[trim={0 0 0 5mm},clip,width=0.875\linewidth]{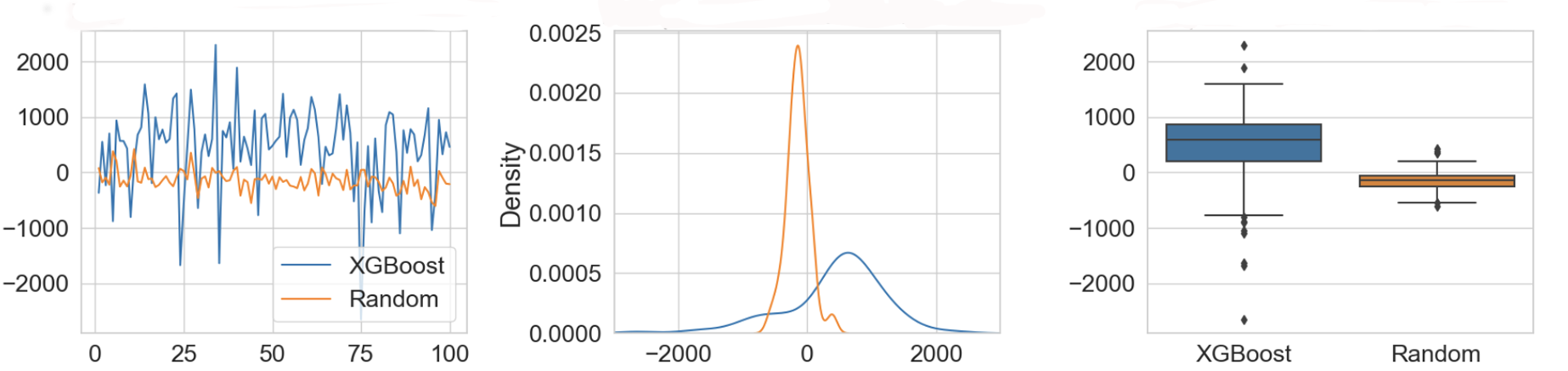}     
\caption{Profit generated from XGBoost compared with Random Betting Agent for Scenario 1.}
\label{fig:4.7}
\end{figure*}

\begin{figure*}
\centering
\includegraphics[trim={0 0 0 6mm},clip,width=0.875\linewidth]{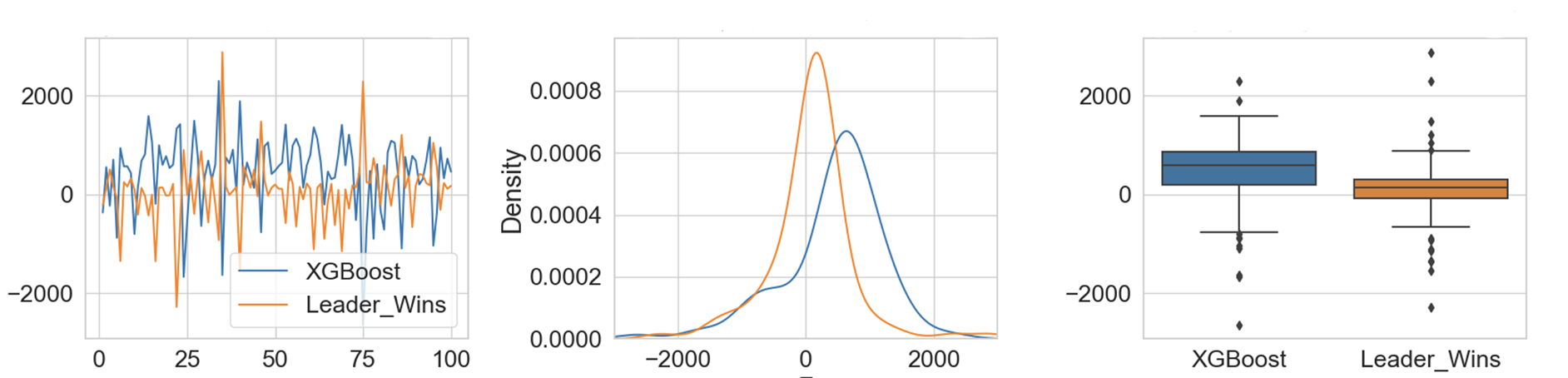}     
\caption{Profit generated from XGBoost compared with Leader Win Agent for Scenario 1.}
\label{fig:4.8}
\end{figure*}

\begin{figure*}
\centering
\includegraphics[trim={0 0 0 8mm},clip,width=0.875\linewidth]{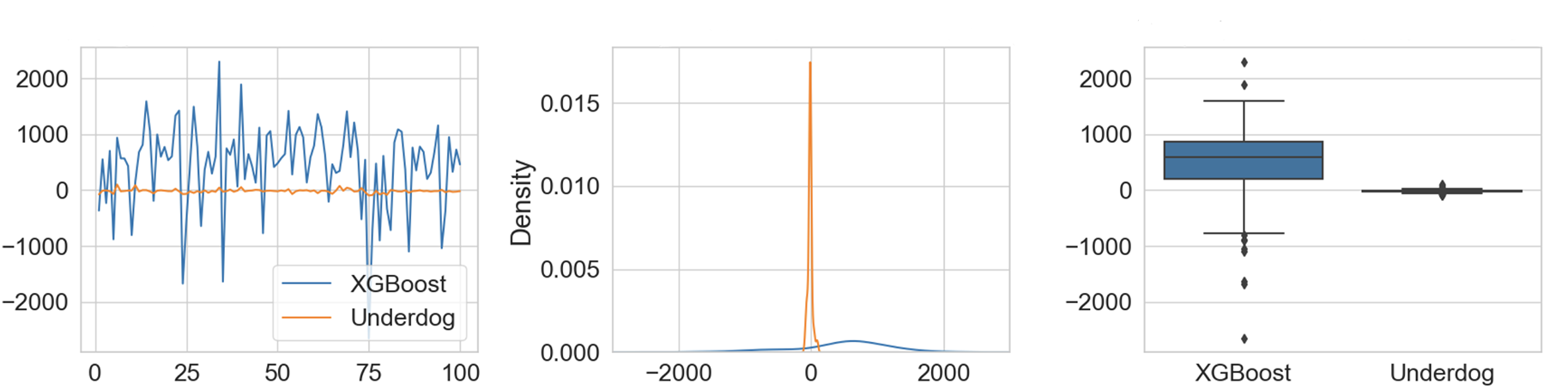}     
\caption{Profit generated from XGBoost compared with Underdog Agent for Scenario 1.}
\label{fig:4.9}
\end{figure*}

\begin{figure*}
\centering
\includegraphics[trim={0 0 0 8mm},clip,width=0.875\linewidth]{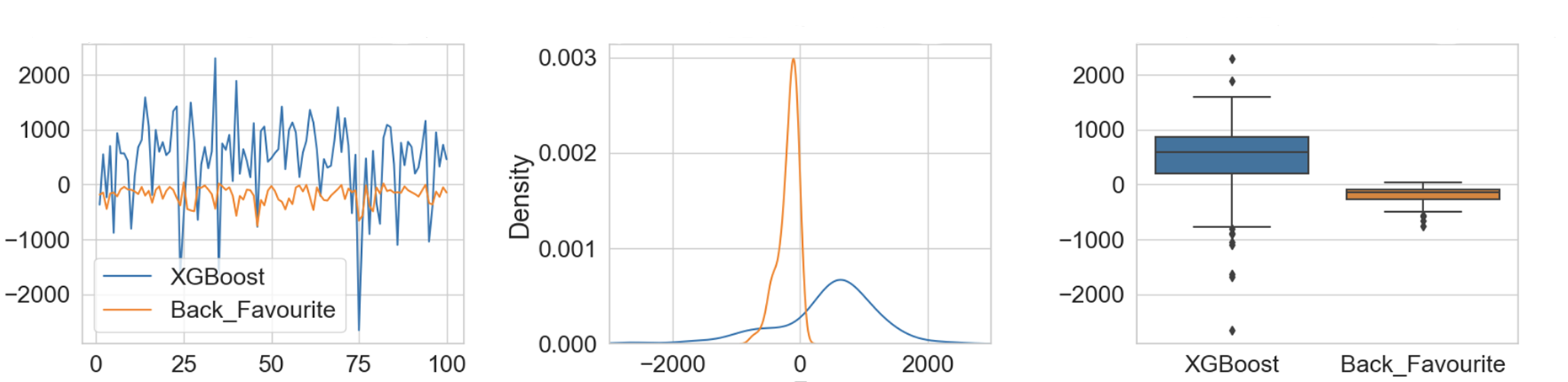}     
\caption{Profit generated from XGBoost compared with Back Favorite Agent for Scenario 1.}
\label{fig:4.10}
\end{figure*}

\begin{figure*}
\centering
\includegraphics[trim={0 0 0 8mm},clip,width=0.875\linewidth]{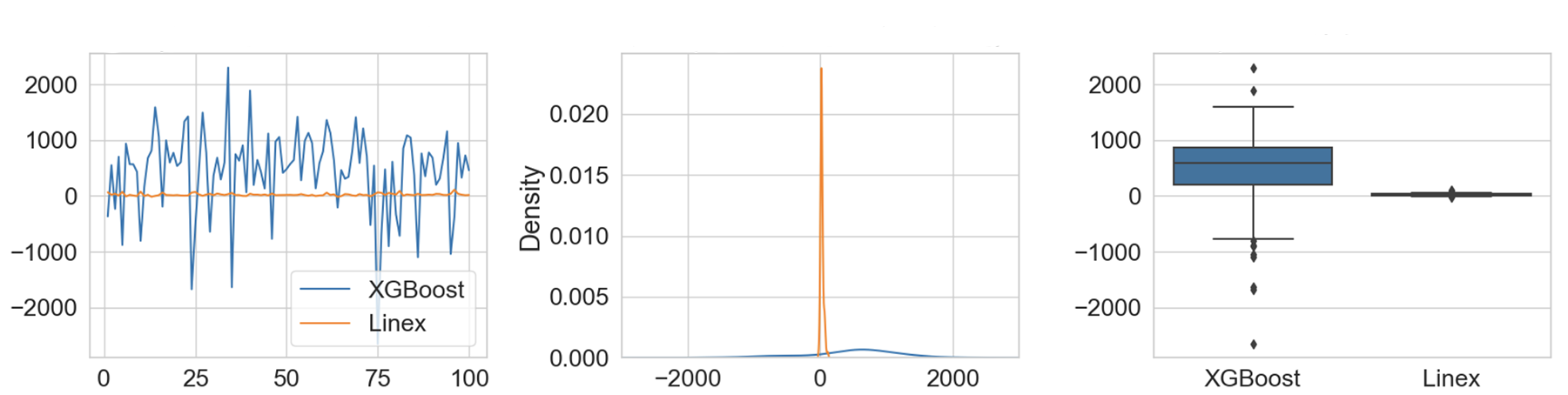}     
\caption{Profit generated from XGBoost compared with Linex Agent for Scenario 1.}
\label{fig:4.11}
\end{figure*}

\begin{figure*}
\centering
\includegraphics[trim={0 0 0 8mm},clip,width=0.875\linewidth]{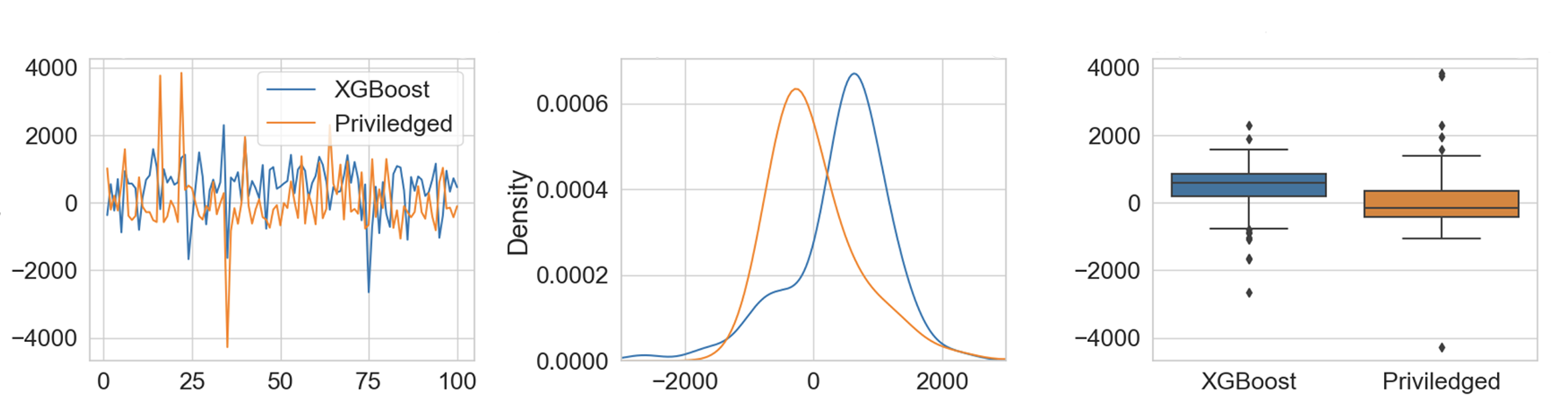}     
\caption{Profit generated from XGBoost compared with Privileged Agent for Scenario 1.}
\label{fig:4.12}
\end{figure*}

In Figures~\ref{fig:4.7} through~\ref{fig:4.12}, a consistent layout is used: the leftmost plot shows the average profit time-series comparison between XGBoost and its counterpart
agent; the central plot exhibits the Kernel Density Estimation (KDE) contrasting XGBoost and the
corresponding agent; the rightmost plot shows a box plot of this comparative data.

{\bf Statistical Examination:} Visual inspection of the Kernel Density
Estimation (KDE) plots of Figures~\ref{fig:4.7} through~\ref{fig:4.12}, indicated that the data distributions were non-Normal, and in each case when we applied the Shapiro-Wilks test, the test outcome confirmed non-Normality. This prompted us to 
use the Wilcoxon-Mann-Whitney U-Test to with null hypothesis that there’s no difference in the profit
averages between the XGBoost agent and other agents, and alternate hypothesis that the XGBoost agent is more profitable. In all cases the null hypothesis is roundly rejected.


\subsubsection{Scenario 2}

{\bf Simulation Setup:}
Retaining the simulation sessions at 100 rounds, a different composition of agents was employed: 5 agents
each for Random, Leader Wins, Underdog, Back Favourite, Linex, XGBoost, and Privilege. 

\begin{figure*} 
\centering
\includegraphics[trim={0 0 0 5mm},clip,width=0.875\linewidth]{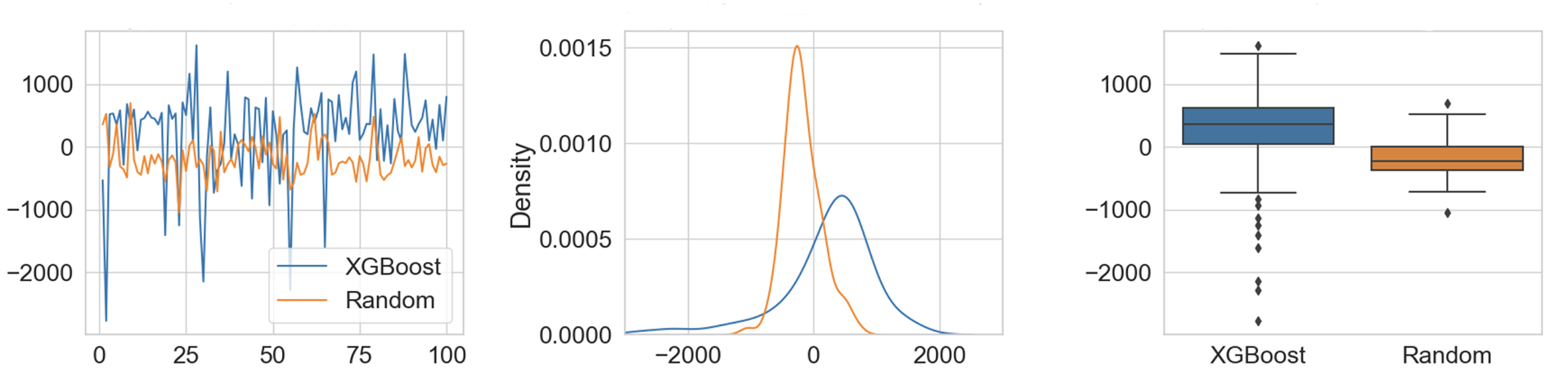}     
\caption{Profit generated from XGBoost compared with Random Betting Agent for Scenario 2.}
\label{fig:4.13}
\end{figure*}

\begin{figure*} 
\centering
\includegraphics[trim={0 0 0 8mm},clip,width=0.875\linewidth]{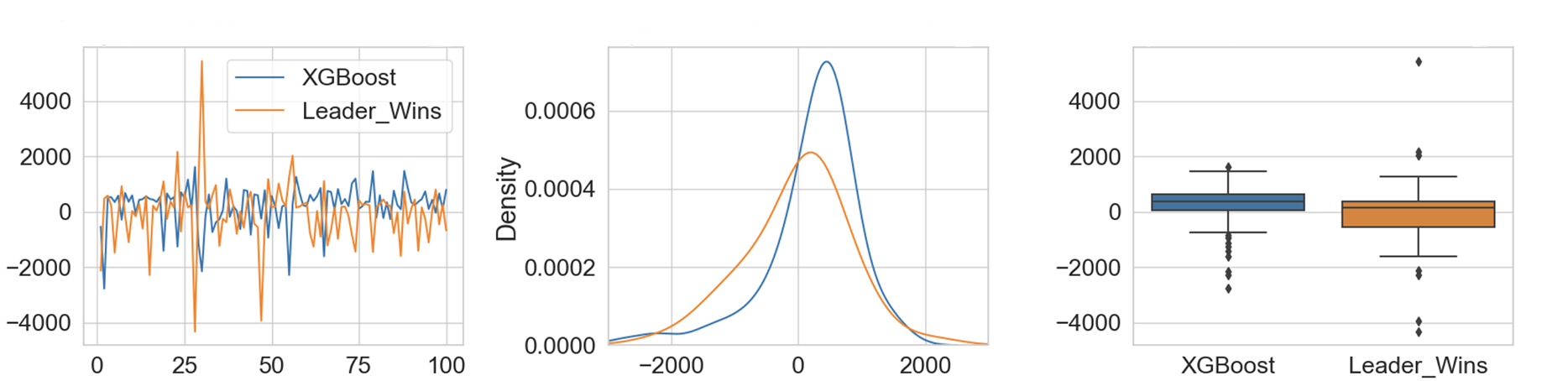}     
\caption{Profit generated from XGBoost compared with Leader Win Agent for Scenario 2.}
\label{fig:4.14}
\end{figure*}

\begin{figure*} 
\centering
\includegraphics[trim={0 0 0 8mm},clip,width=0.875\linewidth]{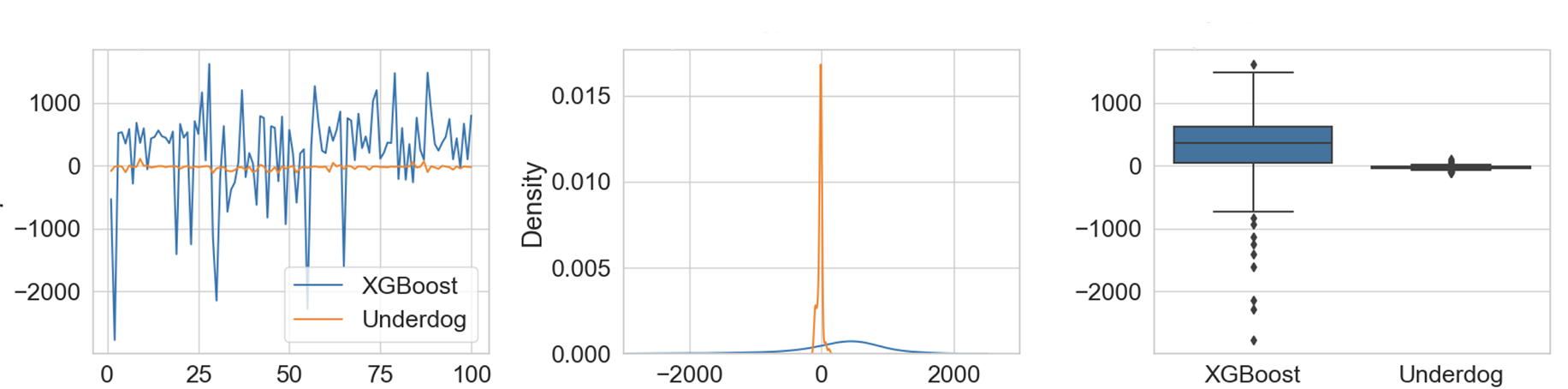}     
\caption{Profit generated from XGBoost compared with Underdog Agent for Scenario 2.}
\label{fig:4.15}
\end{figure*}

\begin{figure*} 
\centering
\includegraphics[trim={0 0 0 8mm},clip,width=0.875\linewidth]{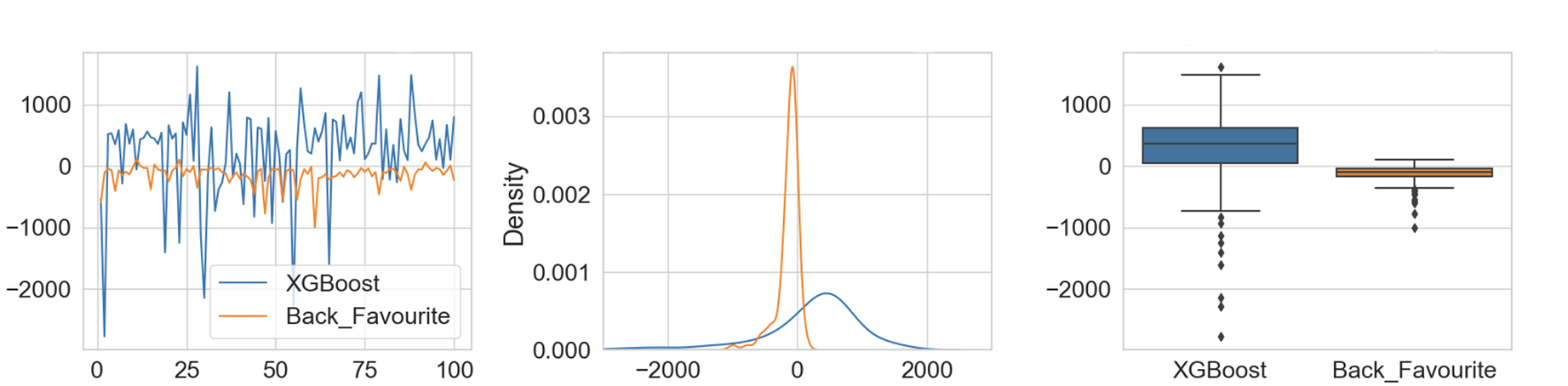}     
\caption{Profit generated from XGBoost compared with Back Favorite Agent for Scenario 2.}
\label{fig:4.16}
\end{figure*}

\begin{figure*} 
\centering
\includegraphics[trim={0 0 0 8mm},clip,width=0.875\linewidth]{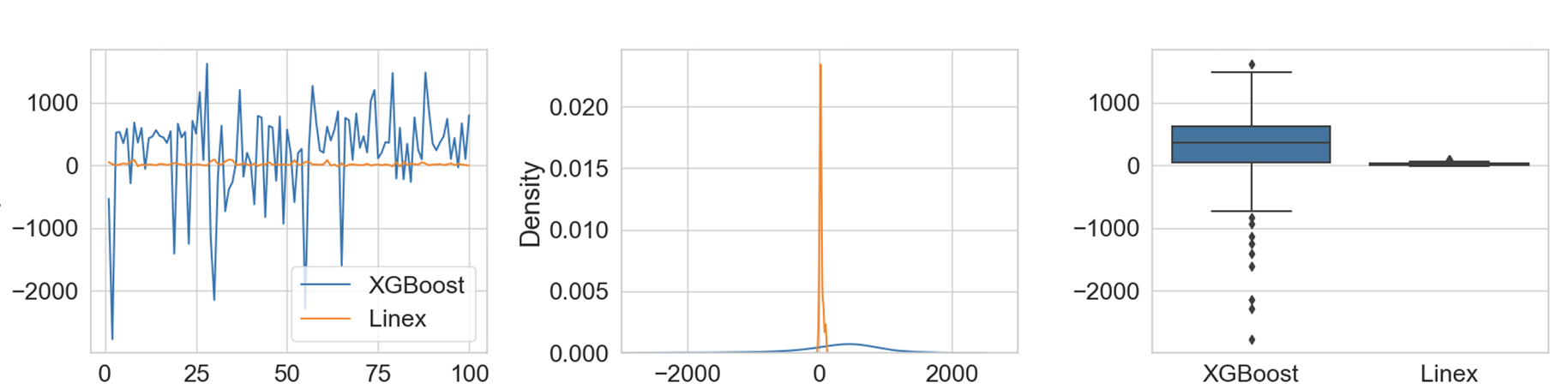}     
\caption{Profit generated from XGBoost compared with Linex Agent for Scenario 2.}
\label{fig:4.17}
\end{figure*}

\begin{figure*} 
\centering
\includegraphics[trim={0 0 0 7mm},clip,width=0.875\linewidth]{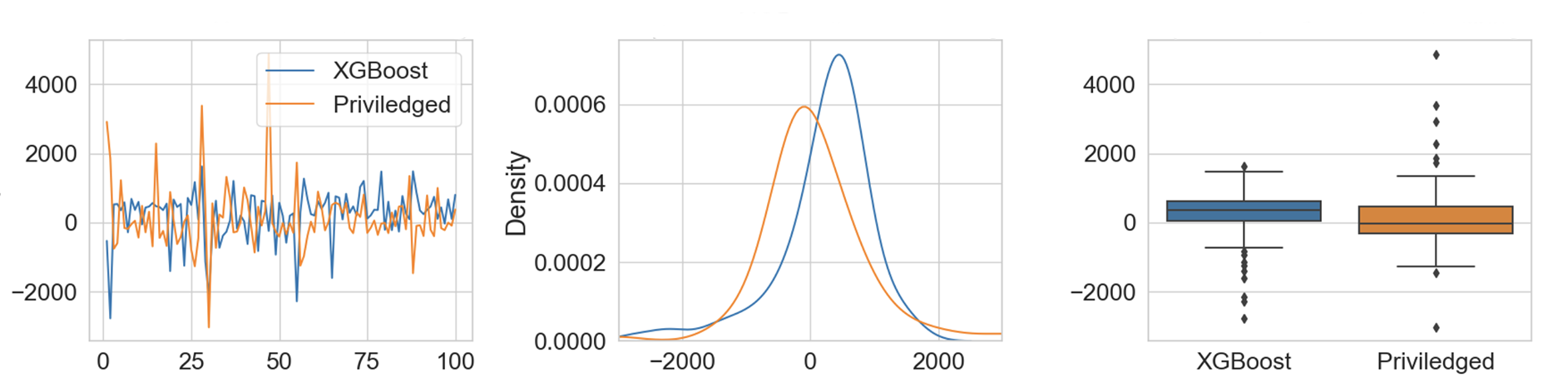}     
\caption{Profit generated from XGBoost compared with Privileged Agent for Scenario 2.}
\label{fig:4.18}
\end{figure*}

{\bf Statistical Examination:}
Similar to Scenario 1, the data’s non-normal distribution was confirmed in each case by the Shapiro-Wilk test and this is confirmed by visual inspection of the Kernel Density Estimation (KDE) plots of Figures~\ref{fig:4.13} through~\ref{fig:4.18}. Consequently, the Wilcoxon-Mann-Whitney U test was used, and in each case the results from the U-test led to the rejection of the null hypothesis (the largest $p$ value, for Privileged/XGBoost, was $0.0017$), further emphasising the XGBoost agent’s performance.

Furthermore, from examination of the plots of Figures~\ref{fig:4.13} through~\ref{fig:4.18} it’s also evident that
the XGBoost betting agent consistently outperforms its peers, the same as in Scenario 1. The line graph
highlights XGBoost’s superior performance, with its values often trending higher. Similarly, the box plot
emphasizes its strong placement, often residing in the upper range of outcomes.

In conclusion, for both scenarios, the XGBoost betting agent demonstrably outperformed its peers
in terms of profit generation. Given these consistent results across different scenarios, it’s obvious that
the XGBoost agent, as modelled and implemented, offers a notable advantage in the context of this
simulation.


\section{Future Work}
\label{sec:future}

Numerous opportunities and potential areas
remain for further investigation, including:

{\bf Variability of Data Collection:} BBE, being a complex system, offers several scenarios and
parameters setting that can influence outcomes. Each race’s length, the number of competitors, and
the mixture of participating agents all contribute uniquely to the final dataset. Currently, the data
extraction from BBE has been largely uniform. However, by introducing more randomness
or systematically varying these parameters, it’s possible to simulate a broader spectrum of race
scenarios. Gathering data from these diverse conditions would likely provide a dataset with richer
contextual information.

{\bf Feature Engineering:} The model currently relies on four primary features is both a strength, for
simplicity, and a limitation, for depth of insight. While distance, rank, time, and stake are crucial,
there exist other features that might further refine the model’s understanding. For instance, the
rate of change of rank over time, interactions between distance and stake, or even cyclic patterns
in betting behaviour could be potential features. Incorporating such sophisticated features could
refine the model’s decision boundaries and offer more precise predictions.

{\bf Model Optimization and Evaluation Metrics:} The choice of the {\em binary:logistic} objective
function has been pivotal for the model’s current design, aiming for binary classification. However,
XGBoost offers a large number of objective functions and evaluation metrics tailored for different
kinds of predictive tasks. By experimenting with other objectives, such as {\em multi:softmax} for multiclass
problems or {\em reg:squarederror} for regression tasks, new insights or even potential performance
improvements could be achieved. This could also lead to the development of betting agents that
can predict more than just binary outcomes, potentially increasing the versatility of the agent in
different betting scenarios.

{\bf Expanded Testing Scenarios:} Here we have used only two testing scenarios. However,
the dynamic nature of the betting domain suggests the potential benefit of a more comprehensive
evaluation, encompassing a broader spectrum of conditions and parameters. The performance and limitations of
the XGBoost betting agent, in comparison to other available agents in the system, could be further
illuminated under an array of diversified scenarios.

For instance, the impact of varying the number of competitors in a race could provide insights
into the agent’s robustness across different competitive landscapes. Similarly, the distance of races
can influence outcome predictability, with certain agents potentially excelling in short sprints while
others might have an edge in longer, more strategic races.

Furthermore, exploring races with different odds ranges can unveil how well the XGBoost betting
agent navigates between high-risk, high-reward situations versus more conservative betting scenarios. Extending the testing to these more diverse scenarios would offer a richer, more
comprehensive view of the XGBoost betting agent’s capabilities, strengths, and potential areas for
improvement

{\bf Online Learning and Feedback Loop Integration:} The field of online machine learning, where
models learn on the go, adapting to new data as it arrives, offers a chance to improve the static
nature of the current implementation. Instead of periodic manual data extraction and retraining, an
integrated feedback loop would allow the XGBoost agent to continuously refine its strategies during
every BBE session.

\section{Conclusion}
\label{sec:conclusion}

The primary contribution of this paper is the introduction of XGBoost learning to the bettor-agents in the BBE agent-based model, offering the opportunity to use BBE as a synthetic data generator and for XGBoost to then learn profitable betting strategies from the data provided from BBE. 
Comparing the XGBoost-learned betting strategy with the performance of the minimally simple strategies pre-coded into BBE demonstrates that XGBoost does indeed
offer a distinct advantage in adaptively learning in-play betting strategies which are more profitable than any of the strategies that were used to create the training data. This serves as a proof-of-concept and in future work we intend to explore application of the methods described here to automatically learn betting strategies that could be profitable if deployed in betting on real-world races.

\bibliographystyle{apalike}
{\small
\bibliography{../../dc_bibliography}}

\vspace*{2cm}
\section*{\uppercase{Appendix: GitHub Repos}}

\label{sec:repos}

The integration of the XGBoost machine learning algorithm into BBE has been separated
into two GitHub repositories, both of which are freely available as open-source Python code from: {\tt https://github.com/ChawinT/}

\subsection*{Synthetic Data Generator}
GitHub repo: {\tt XGBoost\_TBBE/tree/main}
\begin{enumerate}

\item Data Collection: Enhancements were made to the Bristol Betting Exchange (BBE) to facilitate
an efficient data acquisition process.

\item XGBoost Betting Agent: A new component, named the XGBoost betting agent, was introduced
within the betting agent.py file. This serves as a blueprint for embedding machine learning
capabilities into BBE.

\item Model Configuration: The model.json file encapsulates the trained XGBoost model utilized by
the agent for bet predictions.

\end{enumerate}

This repository lays the foundation for data collection and demonstrates a practical blueprint for integrating
machine learning models into the BBE system.

\subsection*{Model Training \& Validation}

GitHub repo: {\tt XGBoost\_ModelTraining}
\begin{enumerate}

\item Model Training: Comprehensive training of the XGBoost model has been conducted, supplemented
with optimization techniques and visualization tools.

\item Statistical Hypothesis Testing: Dedicated sections have been allocated for rigorous statistical
hypothesis testing to validate the reliability of the results.

\end{enumerate}

Together, these repositories form a comprehensive suite for introducing and harnessing the power of
machine learning, specifically XGBoost, in the realm of betting on the BBE platform.

\end{document}